\documentclass[acmtog,authorversion]{acmart}

\usepackage{siunitx}
\usepackage{bm}

\usepackage[ruled]{algorithm2e}

\SetAlFnt{\small}
\SetAlCapFnt{\small}
\SetAlCapNameFnt{\small}
\SetAlCapHSkip{0pt}

\usepackage[capitalize]{cleveref}
\crefname{section}{Sec.}{Secs.}
\Crefname{section}{Section}{Sections}
\Crefname{table}{Table}{Tables}
\crefname{table}{Tab.}{Tabs.}

\setcopyright{acmlicensed}
\acmJournal{TOG}
\acmYear{2024}
\acmVolume{43}
\acmNumber{4}
\acmArticle{151}
\acmMonth{7}
\acmDOI{10.1145/3658196}

\usepackage{epsfig,graphics} 
\usepackage{mathrsfs} 
\usepackage{mathtools} 
\usepackage{esdiff} 
\usepackage{nicefrac,xfrac} 
\usepackage{bm} 
\usepackage{xcolor} 
\usepackage{paralist} 

\newcommand{\etal}{\textit{et al}. }

\DeclareMathOperator{\loss}{\text{Loss}}

\usepackage{amsmath}
\begin{document}
\title{Cricket: A Self-Powered Chirping Pixel}

\author{Shree K. Nayar}
\orcid{0000-0002-6452-6998}
\affiliation{%
\institution{Computer Science Department, Columbia University}
\streetaddress{500 West 120th Street}
\city{New York}
\state{NY}
\postcode{10027}
\country{USA}}
\email{nayar@cs.columbia.edu}

\author{Jeremy Klotz}
\orcid{0009-0005-9408-3541}
\affiliation{%
\institution{Computer Science Department, Columbia University}
\streetaddress{500 West 120th Street}
\city{New York}
\state{NY}
\postcode{10027}
\country{USA}}
\email{jklotz@cs.columbia.edu}

\author{Nikhil Nanda}
\orcid{0009-0002-2145-9710}
\affiliation{%
\institution{Computer Science Department, Columbia University}
\streetaddress{500 West 120th Street}
\city{New York}
\state{NY}
\postcode{10027}
\country{USA}}
\email{nn2522@columbia.edu}

\author{Mikhail Fridberg}
\orcid{0009-0009-1479-4204}
\affiliation{%
\institution{ADSP Consulting, LLC}
\streetaddress{15 Lyndon Road}
\city{Sharon}
\state{MA}
\postcode{02067}
\country{USA}}
\email{fridberg@adspconsulting.com}

\begin{abstract}
We present a sensor that can measure light and wirelessly communicate the measurement, without the need for an external power source or a battery. Our sensor, called cricket, harvests energy from incident light. It is asleep for most of the time and transmits a short and strong radio frequency chirp when its harvested energy reaches a specific level. The carrier frequency of each cricket is fixed and reveals its identity, and the duration between consecutive chirps is a measure of the incident light level. We have characterized the radiometric response function, signal-to-noise ratio and dynamic range of cricket. We have experimentally verified that cricket can be miniaturized at the expense of increasing the duration between chirps. We show that a cube with a cricket on each of its sides can be used to estimate the centroid of any complex illumination, which has value in applications such as solar tracking. We also demonstrate the use of crickets for creating untethered sensor arrays that can produce video and control lighting for energy conservation. Finally, we modified cricket’s circuit to develop battery-free electronic sunglasses that can instantly adapt to environmental illumination.

\end{abstract}

\vspace{-0.2in}
\vspace{-0.2in}

\vspace{-0.2in}
\vspace{-0.2in}
\newcommand{\fid}{\ensuremath{f_{id}}}
\newcommand{\fc}{\ensuremath{f_c}}

\newcommand{\Rone}{\ensuremath{R_1}}
\newcommand{\Rtwo}{\ensuremath{R_2}}
\newcommand{\Rthree}{\ensuremath{R_3}}
\newcommand{\Rfour}{\ensuremath{R_4}}
\newcommand{\Rfive}{\ensuremath{R_5}}
\newcommand{\Rt}{\ensuremath{R_t}}
\newcommand{\Rb}{\ensuremath{R_b}}
\newcommand{\Rspeed}{\ensuremath{R_s}}
\newcommand{\Rgain}{\ensuremath{R_g}}

\newcommand{\Cone}{\ensuremath{C_1}}
\newcommand{\Ctwo}{\ensuremath{C_2}}
\newcommand{\Cthree}{\ensuremath{C_3}}
\newcommand{\Cfour}{\ensuremath{C_4}}
\newcommand{\Cfive}{\ensuremath{C_5}}
\newcommand{\Cspeed}{\ensuremath{C_s}}

\newcommand{\Vone}{\ensuremath{V_1}}
\newcommand{\Vtwo}{\ensuremath{V_2}}
\newcommand{\Vc}{\ensuremath{V_c}}
\newcommand{\Vo}{\ensuremath{V_o}}
\newcommand{\Vf}{\ensuremath{V_f}}
\newcommand{\Vr}{\ensuremath{V_r}}
\newcommand{\Vthree}{\ensuremath{V_3}}

\newcommand{\Sw}{\ensuremath{S}}
\newcommand{\microsec}{$\mu sec$}

\newcommand{\none}{\ensuremath{\bm{\hat{n}_{1}}}}
\newcommand{\ntwo}{\ensuremath{\bm{\hat{n}_{2}}}}
\newcommand{\nthree}{\ensuremath{\bm{\hat{n}_{3}}}}
\newcommand{\nfour}{\ensuremath{\bm{\hat{n}_{4}}}}
\newcommand{\nfive}{\ensuremath{\bm{\hat{n}_{5}}}}
\newcommand{\nsix}{\ensuremath{\bm{\hat{n}_{6}}}}

\newcommand{\ssource}{\ensuremath{\bm{\hat{s}}}}
\newcommand{\ksource}{\ensuremath{k}}
\newcommand{\Ssource}{\ensuremath{\bm S}}
\newcommand{\Sres}{\ensuremath{\bm S}_{r}}
\newcommand{\kmsource}{\ensuremath{\bm k_{m}}}
\newcommand{\ki}{\ensuremath{k_{i}}}
\newcommand{\Nmatrix}{\ensuremath{\bm N}}
\newcommand{\Scent}{\ensuremath{\bm S_{c}}}

\newcommand{\ssixone}{\ensuremath{\bm{\hat{n}_{6_{1}}}}}
\newcommand{\ssixtwo}{\ensuremath{\bm{\hat{n}_{6_{2}}}}}
\newcommand{\ssixi}{\ensuremath{\bm{\hat{s}_{6_{i}}}}}
\newcommand{\scent}{\ensuremath{\bm{\hat{s}_{c}}}}

\newcommand{\Sone}{\ensuremath{\bm S_{1}}}
\newcommand{\Stwo}{\ensuremath{\bm S_{2}}}
\newcommand{\Sthree}{\ensuremath{\bm S_{3}}}
\newcommand{\Sfour}{\ensuremath{\bm S_{4}}}
\newcommand{\Sfive}{\ensuremath{\bm S_{5}}}
\newcommand{\Ssix}{\ensuremath{\bm S_{6}}}
\newcommand{\Sseven}{\ensuremath{\bm S_{7}}}
\newcommand{\Seight}{\ensuremath{\bm S_{8}}}
\newcommand{\Si}{\ensuremath{\bm S_{i}}}

\newcommand{\Ivect}{\ensuremath{\bm I}}
\newcommand{\Nvect}{\ensuremath{\bm N}}
\newcommand{\Ione}{\ensuremath{I_1}}
\newcommand{\Itwo}{\ensuremath{I_2}}
\newcommand{\Ithree}{\ensuremath{I_3}}
\newcommand{\Ifour}{\ensuremath{I_4}}
\newcommand{\Ifive}{\ensuremath{I_5}}
\newcommand{\Isix}{\ensuremath{I_6}}

\newcommand{\Lampsize}{\ensuremath{\bm L}_{n \times 1}}
\newcommand{\Meassize}{\ensuremath{\bm M}_{m \times 1}}
\newcommand{\Transsize}{\ensuremath{\bm K}_{m \times n}}
\newcommand{\Lamp}{\ensuremath{\bm L}}
\newcommand{\Meas}{\ensuremath{\bm M}}
\newcommand{\Trans}{\ensuremath{\bm K}}
\newcommand{\Amb}{\ensuremath{\bm A}}
\newcommand{\Lampnow}{\ensuremath{{\bm L}(t)}}
\newcommand{\Measnow}{\ensuremath{{\bm M}(t)}}
\newcommand{\Measnext}{\ensuremath{{\bm M}(t+1)}}
\newcommand{\Measdes}{\ensuremath{{\bm M}_D}}
\newcommand{\Lampnext}{\ensuremath{{\bm L}(t+1)}}

\begin{teaserfigure}
\includegraphics[width=1.0\linewidth]{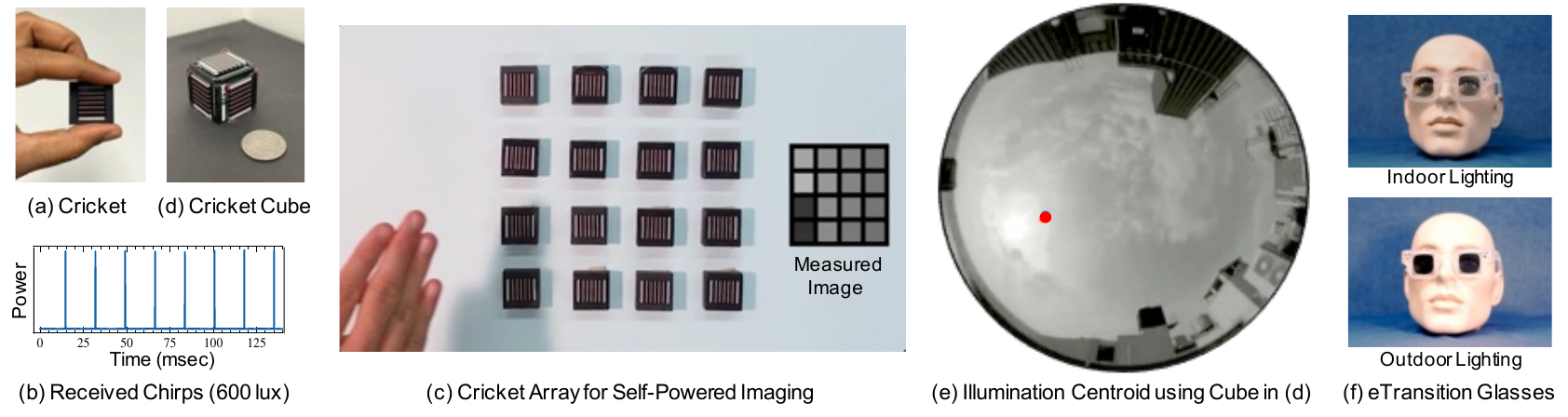}
\vspace{-0.25in}
\caption{{\bf Cricket: An untethered, self-powered light sensor.} (a) Cricket is a battery-free light sensor based on a minimalist analog circuit that emits short but strong radio-frequency (RF) chirps. (b) The duration between consecutive chirps is inversely proportional to the incident light energy. Cricket has a very wide dynamic range -- from $10 \ \unit{lux}$ to $170,000 \ \unit{lux}$, which is well above direct sunlight. (c) Each cricket operates at a fixed RF frequency, and hence a large number of them can be deployed using a small frequency band. Here, an array of crickets is used to emulate a low-resolution video camera. (d) A compact cube with a cricket on each face can be used to precisely measure the centroid of any illumination field, such as the sky in (e). (f) A modified version of the cricket circuit is used to create battery-free transition eyeglasses that can adapt to light in milliseconds, compared to photochromic transition glasses that take roughly half a minute to darken and two minutes to lighten.}
  \label{fig:teaser}
\end{teaserfigure}

\maketitle

\section{Introduction}\label{sec:intro}

Today, light sensors are used in a wide range of applications. In security and monitoring, they are used to detect motion, lighting changes and even simple activities. In safety, they are configured to create light curtains for proximity and intrusion detection. In the context of sustainability, they enable solar panels to track the sun and indoor lighting systems to conserve energy. In the realm of environmental sensing, they help monitor the weather and air quality. In agriculture, they are used to measure the health of crops and control irrigation. Light sensors are also used as light probes in the fields of vision and graphics. Finally, millions of light sensors are used in image sensors to produce photos and videos. 

In many of the above applications, it is highly beneficial to have the light sensors deployed in an environment without requiring them to be tethered. An untethered light sensor should be able to measure incident light and wirelessly communicate the measurement without an external power supply or a battery that needs to be replaced. Our work is focused on the development of such a self-powered (battery-free) sensor that uses light to measure light. 

Our sensor, referred to as “cricket,” is based on a minimalist design and is shown in \cref{fig:teaser}(a). It is a simple analog circuit that is asleep for most of the time. It uses a photovoltaic cell that charges a capacitor. When the capacitor reaches a certain voltage, the circuit wakes up and transmits a short but strong radio-frequency (RF) chirp, which can be received by a radio located in the environment. Each generated chirp entirely depletes the energy harvested by the capacitor, causing the circuit to go back to sleep and the harvesting to start again. Since the time it takes to charge the capacitor is directly proportional to the intensity of the incident light, the duration between two consecutive chirps is a measure of the light intensity (see \cref{fig:teaser}(b)). In other words, the frequency of chirps received by the radio is proportional to the light intensity, and the carrier frequency of each chirp reveals the cricket's identity. 

Since each cricket operates at a fixed carrier frequency, a large number of crickets can be defined in a small frequency band and used simultaneously in an environment. In \cref{fig:teaser}(c), an array of crickets is used to emulate a low-resolution camera that can produce video.

Our cricket has several important characteristics. First, a single photovoltaic cell is used to both measure the incident light and power the circuit. Since we use inter-chirp duration to encode the measured light energy, cricket has an unprecedented dynamic range – from $10 \ \unit{lux}$  to $170,000 \ \unit{lux}$ (well above direct sunlight) in the case of our prototype. Equally important is the fact that a reduction in the size of the photovoltaic cell, and hence the size of the cricket, simply translates to an increase in the duration between chirps. While our prototype uses a photovoltaic with an active area of $16\times16 \ \unit{\square \mm}$, we have experimentally verified the feasibility of using a photovoltaic with a sensing area as small as $1.7 \times 1.7 \ \unit{\square \mm}$. We have extensively evaluated the performance of cricket by measuring its response function, signal-to-noise ratio and dynamic range. 

We introduce the concept of a cricket cube, which has a cricket on each of its six faces (see \cref{fig:teaser}(d)). We show that the six measurements captured by the cube can be used to compute the ``centroid'' of the illumination arriving at a point, irrespective of the complexity of the illumination. Given the wide dynamic range of crickets, the cube is able to accurately estimate illumination centroids for both dimly-lit indoor scenes and outdoor scenes with direct sunlight (see \cref{fig:teaser}(e)). Among other applications, the computed centroid can be used to orient solar panels to optimize the generation of renewable energy.

A popular application of light sensors is in energy conservation, where light measurements are used to control the intensities of lamps in an indoor setting.  Crickets, being untethered, are particularly suited for this application as they are easily deployed in any space. The measurements they produce are used to ensure that the illumination of chosen areas (e.g., desks, workbenches, plants, etc.)  are kept constant, even as the ambient lighting (e.g., sunlight through windows) varies. In another novel application, we have modified the cricket circuit to develop battery-free electronic transition sunglasses (see \cref{fig:teaser}(f)). Traditional transition glasses use photochromic coatings that are extremely slow to adapt to lighting changes. In contrast, our “eTransition” glasses adapt to lighting changes almost instantaneously. 
We conclude the paper with a brief discussion on the limitations of the current prototype and our planned future work.

\section{Related Work}\label{sec:related}

The most closely related work to ours is Sozu, a novel self-powered wireless RF tag~\cite{zhang_sozu_2019} that can be powered by energy harvested from a variety of physical phenomena. The authors have also demonstrated the use of Sozu for the measurement of light. There are two significant differences between our cricket and Sozu. First, in Sozu's case, the transmitted frequency is a function of the light level, requiring each tag to use a range of frequencies for its measurement. In contrast, cricket transmits at a single frequency and uses time between chirps to encode light intensity. This allows us to use a larger number of crickets within any given frequency band. A second difference is that the maximum intensity measured by Sozu is limited by the maximum voltage the energy source (photovoltaic in the case of light) can produce. Since cricket measures the time it takes to harvest a fixed amount of energy, it is not limited by the peak voltage of the photovoltaic and is able to produce light measurements over a very wide dynamic range with high accuracy.

Cricket maps light intensity to time by measuring the time it takes to collect a fixed amount of energy. The concept of mapping light intensity to a time measurement has been explored in the past~\cite{forchheimer-near-sensor-image-processing} and implemented on an image sensor~\cite{brajovic-sorting-image-sensor, culurcielloBiomorphicDigitalImage2003}.
An interesting implementation of this sensing method is the spiking camera~\cite{zhuRetinaInspiredSamplingMethod2019}, an image sensor in which each pixel generates a spike when the accumulated energy reaches a threshold. The spiking camera has been used to demonstrate a variety of vision tasks~\cite{Zhao_2021_ICCV, Zheng_2021_CVPR, Hu_2022_CVPR} and fabricated on a chip~\cite{huang1000FasterCamera2023}.
Like all of the above works, cricket maps the incident light level to a time measurement. However, our key innovation is the use of harvested energy both to measure and wirelessly transmit the incident light intensity without a battery or an external power source.

Several other works have also explored the design and application of self-powered light sensors. 
Zhang \etal~\shortcite{zhang_optosense_2020} proposed OptoSense, a self-powered sensor that can be adhered to surfaces to detect simple activities such as 2D gestures or events such as a door opening. In a similar vein, Zhang \etal~\shortcite{zhang_flexible_2022} proposed a computational sensor that performs simple arthimetic on the outputs of an array of photodiodes in hardware to reduce power consumption. 
Both of the above designs use a photovoltaic panel for energy harvesting, separate photodiodes for sensing, a microcontroller for readout and transmission, and a battery for power when the harvested energy is inadequete. In contrast, our cricket is a simple analog circuit that uses a single photovoltaic to both measure and transmit light levels. Cricket can be viewed as having a minimalist design as it does not use a microcontroller and only transmits measurements when it can ``afford'' to. This approach enables cricket to wirelessly provide measurements at low light levels (down to $10 \ \unit{lux}$) while being compact and fully self-powered.

Power harvesting light sensors have been used to demonstrate a variety of other applications, including lighting control~\cite{pandharipande_light-harvesting_2013} and gesture recognition~\cite{li_self-powered_2018, ma_solargest_2019, sandhu_solar_2021}. 
In these cases, a battery is 
needed for measurement and transmission in low-light conditions.
There is also related work that uses RF backscatter as an alternate source of energy for powering battery-free light sensors ~\cite{varshney_battery-free_2017} and cameras~\cite{naderiparizi_wispcam_2015, naderiparizi_towards_2018, josephson_wireless_2019}. In these cases, the sensor derives it energy from an RF reader (as in the case of RFID tags) and can only provide measurements when powered by the reader. 

The last application we describe is a pair of electronic transition sunglasses that can instantly adapt to environmental illumination without the use of a battery. 
Traditional transition glasses, which are in wide use today, are based on the chemical phenomenon called photochromism~\cite{smith_photochromic_1967, dorion_photochromism_1970}. Photochromism is known to be very slow to respond to changes in light, taking 30-40 seconds to go from light to dark and over 2 minutes to return from dark to light. Recently, a few sunglasses have appeared in the market that use a liquid crystal (LC) light valve in front of each eye~\cite{lavie, deko}. These devices respond instantaneously to incident light but require a battery to power them. Since the first part of our cricket circuit is a self-powered pulse generator, we modified the circuit to change the duty cycle of the pulse generator and have it drive an LC valve. We used this idea to create a pair of transition sunglasses that can function without the use of a battery.

Finally, there is growing interest in developing image sensors that are partially powered by energy harvested from light incident upon the sensor~\cite{fish_self-powered_2005, fish_cmos_2006, shi_novel_2011, ay_cmos_2011, law_low-power_2011}.
Nayar \etal~\shortcite{nayar_towards_2015} demonstrated a self-powered  camera with $30\times 40$ pixels that harvests energy to read out a full image. While this camera can produce video without an external power supply, it does not harvest enough energy to wirelessly transmit its images. 

\vspace{0.05in}
\section{Cricket: A Chirping Pixel}\label{sec:cricket_circuit}

We begin with the reason we refer to our sensors as crickets. In nature, male crickets produce loud chirps that are used as mating calls. The chirp is made using a phenomenon called stridulation where serrations on one wing are rubbed against a scraper on the other wing. Brooks~\shortcite{brooks_influence_1881} discovered a relationship between the number of chirps a cricket makes per unit time and the ambient temperature of its habitat. Later, Dolbear~\shortcite{dolbear_cricket_1897} put forth an empirical model for this relationship: $T_F = 50 + (N_{60}-40)/4$, where $N_{60}$ is the number of chirps in a minute and $T_F$ is the temperature in Fahrenheit. The behavior of our light sensor is analogous, in that, the chirps are short and loud, and the number of chirps per second is proportional to the intensity of incident light. The chirps of our sensor are inaudible as they are in the radio frequency regime. 

\subsection{Circuit}

\begin{figure}
    \centering
    \includegraphics[width=\linewidth]{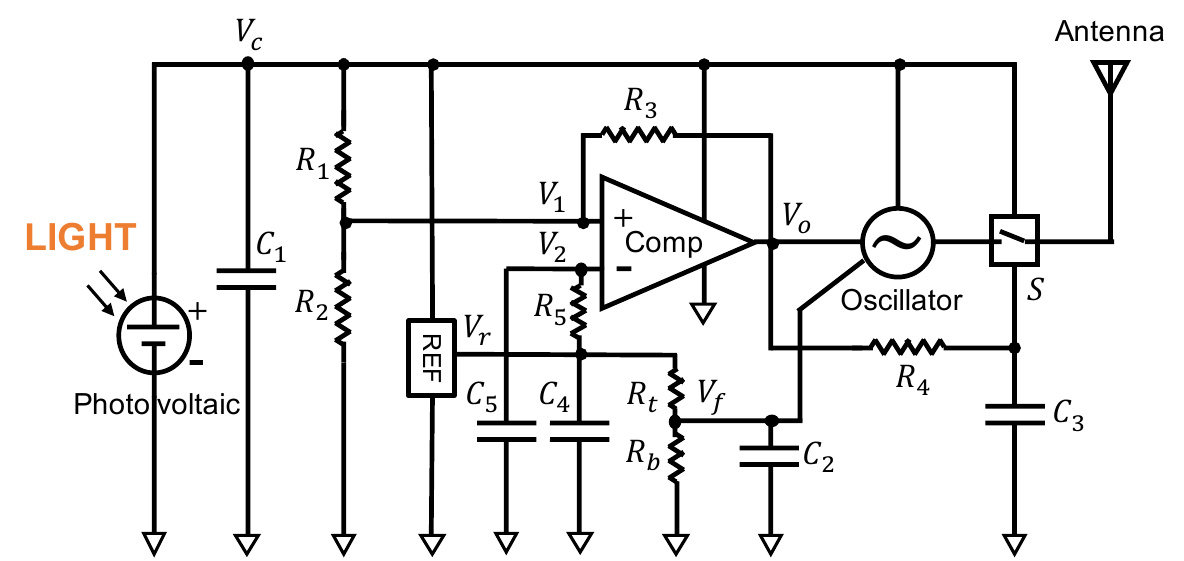}
    \caption{{\bf Cricket circuit.} When the photovoltaic cell is exposed to light, the voltage across the capacitor \Cone \ begins to rise. At a certain voltage, the circuit comes alive and the oscillator generates a short and strong chirp. The time between consecutive chirps reveals the intensity of incident light.}
    \label{fig:cricket-circuit}
\end{figure}

The circuit of a cricket is shown in \cref{fig:cricket-circuit}. When the photovoltaic cell is exposed to light it charges capacitor \Cone, and voltage \Vc\ begins to rise. At some value of \Vc, the comparator comes to life. At that point, the output of the comparator is $0$. At an even higher value of \Vc, the output of the comparator jumps from $0$ to \Vc, the oscillator is activated and generates
an RF frequency (in the GHz range).\footnote{We designed our crickets to function in the GHz range to keep the size of the antenna small. Our power calculations indicate that our current prototype could be an FCC compliant Part 15 transmitter.  For applications where a larger antenna can be used and transmission over longer ranges and better penetration through dense media is required, the cricket can be configured to function in the MHz range. This is done by simply changing the oscillator, the antenna and the values of a few of the passive components.}  The carrier frequency \fid\ of the oscillator, which is the identifier of the cricket, is determined by the voltage \Vf. In our implementation, \fid\ is preset by selecting the resistor \Rb. 

Although the oscillator is now active, it is not yet connected to the antenna due to the switch \Sw. The closing of this switch is delayed to allow the preset voltage \Vf\ to stabilize and ensure that the frequency applied to the antenna is precise and stable. In our current implementation, this delay has been set to roughly $10 \ \unit{\micro s}$. After the delay, the oscillator is connected to the antenna, which transmits an RF chirp for about $30 \ \unit{\micro s}$.\footnote{Strictly speaking, a chirp is defined as a signal in which the frequency changes. In our case, the frequency does change as the oscillator settles, but we only use the part of the signal for which the frequency is constant.} This chirp duration is limited solely by the fact that the oscillator is, by far, the highest power consumer in the circuit. Hence, while the oscillator is active, it drains all the energy stored in \Cone. As a result, \Vc\ falls rapidly until the comparator output goes to $0,$ and the oscillator shuts down. At this point, \Cone\ begins to recharge, and \Vc\ rises again. 

\begin{figure}
    \centering
    \includegraphics[width=0.99\linewidth]{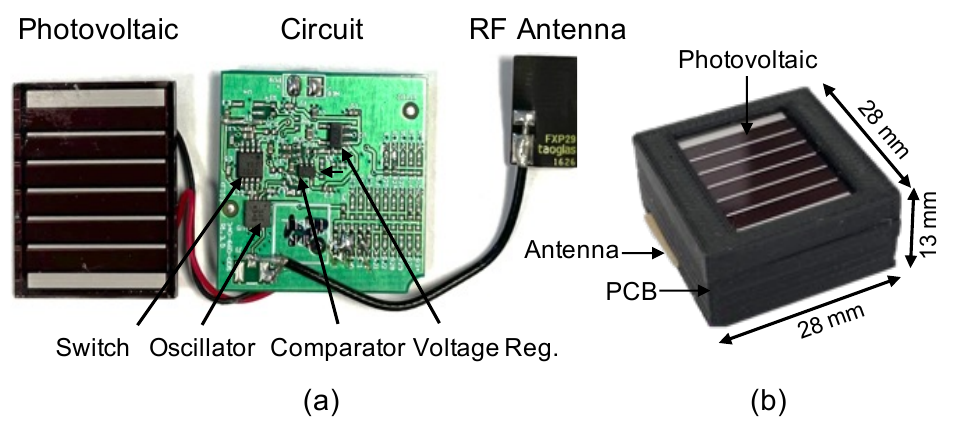}
    \vspace{-0.1in}
    \caption{{\bf Cricket prototype.} The photovoltaic cell, printed circuit board and RF antenna are stacked and packaged in a compact 3D-printed case.}
    \label{fig:cricket-pcb}
      \end{figure}
      
\begin{figure}
    \centering
    \includegraphics[width=0.94\linewidth]{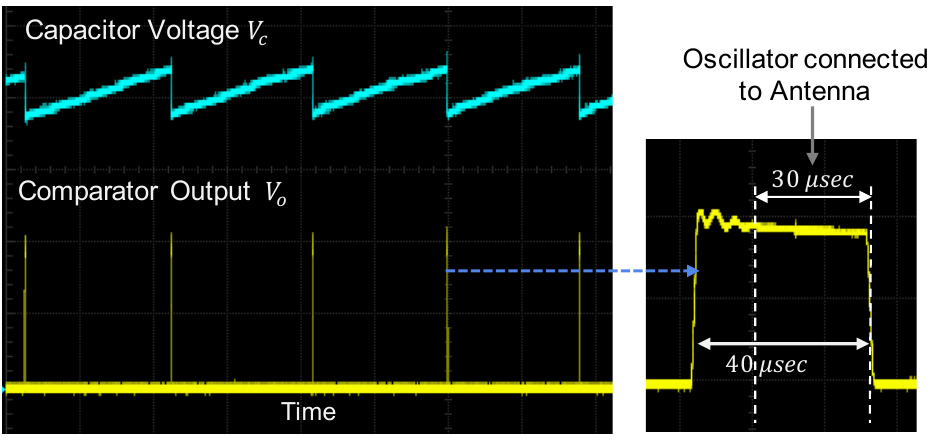}
    \caption{{\bf Cricket voltages.} The waveforms of the voltage \Vc\ of the capacitor \Cone\ and the comparator output \Vo, measured using an oscilloscope. The oscillator is connected to the antenna only after its output frequency has stabilized.}
    \label{fig:voltages}
\end{figure}

The duration between chirps is determined by the time it takes for \Cone\ to recharge, which in turn depends on the intensity of incident light. {\em In short, the carrier frequency \fid\ of each chirp represents the identity of the cricket, and the frequency \fc\ of the chirps (chirps per second) is proportional to intensity of the incident light.}

In the supplemental material, we have provided a more detailed description of how the circuit works as well as the values and model numbers of the components we used in our current prototype.

\Cref{fig:cricket-pcb}(a) shows the key components of the cricket, and \cref{fig:cricket-pcb}(b) shows the final packaged prototype which was used in all our experiments. The layout of the printed circuit board is available online~\cite{cricket-website}. \Cref{fig:voltages} shows the voltage \Vc\ across \Cone\ and the comparator output voltage \Vo\ measured using an oscilloscope, for a light level of $150 \ \unit{lux}$. Note the sawtooth waveform of \Vc, the period of which is inversely proportional to the incident light level. The waveform of the comparator output \Vo\ is a train of short pulses, one of which is magnified. The shape of this pulse is unaffected by the incident light level. It is roughly $40 \ \unit{\micro s}$ wide, during which time the oscillator is active.  Due to the delay discussed previously, it is only for the last $30 \ \unit{\micro s}$ of the pulse that the oscillator is connected to the antenna. 

When the oscillator is off, the circuit only consumes about $2.2 \ \unit{\uW}$. 
Virtually all the power is used by the oscillator when it is on, which is about $10 \ \unit{\mA}$ at $2.9 \ \unit{\V}$ for $40 \ \unit{\micro s}$. We have estimated that a redesign of the oscillator would enable it to generate chirps of the same power while consuming roughly $1 / {10}^{\text{th}}$ the power consumed by the current oscillator. This gain can be used to reduce the area of the photovoltaic cell by a factor of 10. Alternatively, an oscillator that is $10$ times more efficient can be used to increase the transmittance range of cricket by  a factor of $\sqrt{10} = 3.33$.

\subsection{Signal Processing: Chirp Detection}
\begin{figure}
    \centering
    \includegraphics[width=0.98\linewidth]{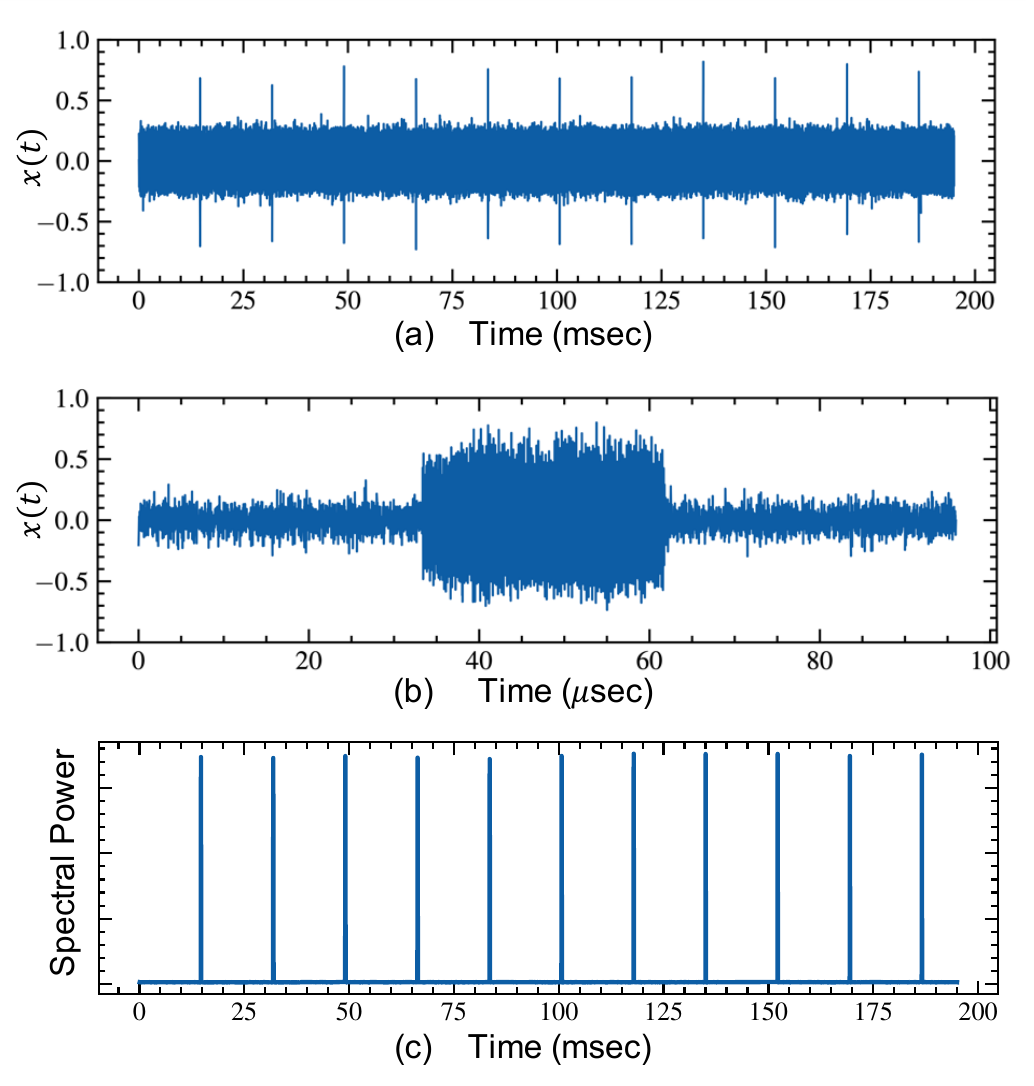}
    \vspace{-0.15in}
    \caption{{\bf Chirp detection.} (a) Signal from a single cricket received by the software defined radio. (b) A time-expanded view of a single chirp. (c) The known carrier frequency \fid \, of the cricket is used to detect chirps in frequency domain. Here, the power within a small window around \fid \, is plotted as a function of time. The frequency \fc \, of received chirps (chirps per second) is proportional to the incident light level.}
    \label{fig:signal-processing}
\end{figure}

As mentioned above, each cricket is assigned a specific carrier frequency $\fid$. We have designed all our crickets to lie in the range $2.04\ \unit{\GHz} \le \fid \le 2.11\ \unit{\GHz}$. The chirps are received by a software-defined radio (SDR) located in the environment, which is tuned to the middle of the above range and reads a $50\ \unit{\MHz}$ baseband signal $x(t)$, an example of which is shown in \cref{fig:signal-processing}(a). We found that with line of sight, the crickets could be placed up to 40 feet away from the SDR for it to receive strong chirps. The spikes seen in \cref{fig:signal-processing}(a) correspond to chirps received by the SDR, one of which is magnified in \cref{fig:signal-processing}(b). Since we know \fid\ for each cricket, we detect its chirps in the frequency domain. First, we compute a $4096$-point ($80 \ \unit{\micro s}$) discrete Fourier transform (DFT) with a Hamming window in $2048$-point ($40 \ \unit{\micro s}$) steps. Since a cricket's \fid\ may vary slightly with temperature and humidity, we search for its chirps within a $2 \ \unit{\MHz}$ band centered around \fid. In each $4096$-point window, we compute the average power within this $2 \ \unit{\MHz}$ band, as shown in \cref{fig:signal-processing}(c). A chirp is detected when the power exceeds a threshold. 

Note that the time resolution of each detection depends on the number of samples by which the DFT window is shifted. While we have typically used a shift of $2048$ samples, in some applications we have used smaller shifts to achieve higher time resolution. 
Our signal processing code is available online~\cite{cricket-website}.

\subsection{Performance}\label{character}

\begin{figure*}[t] 
\centering 
\includegraphics[width=1.0\linewidth]{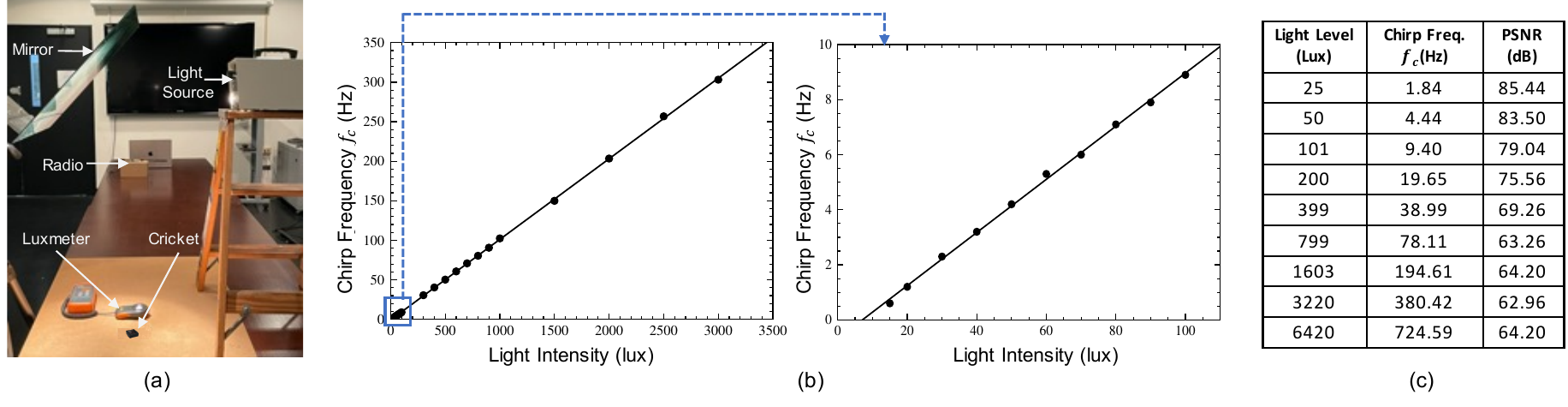}
\vspace{-0.2in}
\caption{{\bf Performance characteristics of crickets.} (a) Experimental setup used to evaluate the performance of crickets. The cricket is illuminated using a regulated halogen light source. A luxmeter is placed close to the cricket to obtain ground truth light levels. In these experiments, the software defined radio was placed roughly 12 ft from the cricket. (b) The radiometric response of a cricket was found to be linear. 
Also shown is a zoomed-in view of the response function for low light levels ($0-100 \ \unit{lux}$). Ideally, the response function should pass through the origin. It does not in our measurements because of a small error (roughly $7 \ \unit{lux}$) in the measurements produced by the luxmeter, which were used as ground truth. (c) The PSNR of a cricket for different light levels.}
\label{fig:radiometry}
\end{figure*}

We have conducted a detailed analysis of the performance of our cricket prototype. \Cref{fig:radiometry}(a) shows the setup we used for most of our experiments, which includes a regulated halogen light source, a mirror to control the distance between the source and the cricket, a luxmeter placed next to the cricket to obtain ground truth light levels, and a software radio placed about 12 feet from the cricket. 
\vspace{0.1in}
\noindent
{\bf Radiometric Response Function:}
To characterize the cricket's response function, we measured the frequency \fc\ of detected chirps for light levels varying from $10 \ \unit{lux}$ to $3500 \ \unit{lux}$, which was the maximum light level we could generate using our regulated source. The measured response function is shown in \cref{fig:radiometry}(b) and is seen to be linear. 
In order to illustrate that the linearity of the response function holds for low light levels, we also show a zoomed-in view of the lower end of the response function (blue box) which includes measurements in the range $0 - 100 \ \unit{lux}$.  The best line that fits our measurements shows that the response remains linear in this range. Ideally, this line should pass through the origin, but it intersects the $x$-axis at about $7 \, \unit{lux}$. We attribute this offset to a bias in our luxmeter readings which were used as our ground truth light measurements. 

We have fabricated a total of $24$ crickets (see \cref{fig:time-varying}(a)), all of which have linear response functions with slightly different gains due to expected variability in the values of the components used. 
Each cricket produces roughly $0.1$ chirps per lux, per second.

\vspace{0.1in}
\noindent
{\bf Dynamic Range:}
Cricket has a remarkably wide dynamic range as it can measure light levels ranging from $10 \ \unit{lux}$ (dimly lit room) to 170,000  $\unit{lux}$, which is well above direct sunlight.\footnote{Direct sunlight on a clear day is estimated to be around 130,000 $\unit{lux}$ \cite{sunlight-bright}.} The lower limit of $10 \ \unit{lux}$ is due to the fact that the oscillator we used requires $2.9 \ \unit{\volt}$ to chirp, and the photovoltaic we used does not produce that voltage for light levels less than $10 \ \unit{lux}$. Note that this is not a fundamental limit -- this lower limit can be significantly lowered by custom-designing the oscillator and the photovoltaic.
The upper limit of the dynamic range results from the fact that, at some light level, the received chirps, which are roughly $30 \ \unit{\micro s}$ wide, will abut each other, making them undetectable as distinct chirps. To estimate the upper limit, we take a conservative approach and require consecutive chirps to be separated by at least $30 \ \unit{\micro s}$ to be reliably detected. This condition corresponds to a light level of approximately 170,000 $\unit{lux}$. Therefore, we can approximate the dynamic range as:
\begin{align}
    20 \log_{10} \left( \frac{170,000\ \unit{lux}}{10\ \unit{lux}} \right) \ \unit{\decibel}
    = 84.6 \ \unit{\decibel}.
\end{align}
In our experiments, we have measured light levels as low as $10 \ \unit{lux}$ in a dark room and as high as 62,800 $\unit{lux}$ under sunlight (see \cref{fig:centroid}).

\begin{figure}[h] 
\centering 
\includegraphics[width=0.9\linewidth]{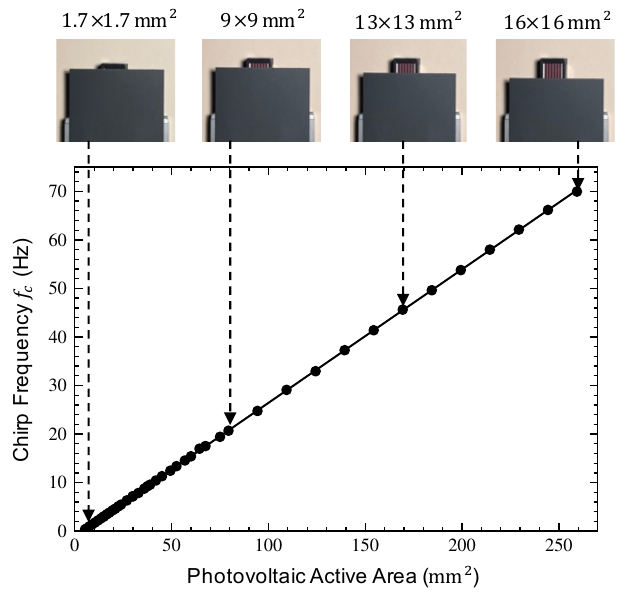}
\vspace{-0.15in}
\caption{{\bf Miniaturization of cricket.} We varied the active area of the photovoltaic by sliding a mask over a cricket. The frequency \fc\ of chirps decreases linearly with the active area, for any given light level. We found that the active area of the current prototype could be reduced by a factor of $100\times$, which suggests that by trading-off the number of chirps received per second, the active area of a cricket can be as small as $1.7 \times 1.7 \ \unit{\square \mm}$.}
\vspace{-0.15in}
\label{fig:cricket-size}
\end{figure}

\begin{figure*}[t]
\includegraphics[width=1.0\linewidth]{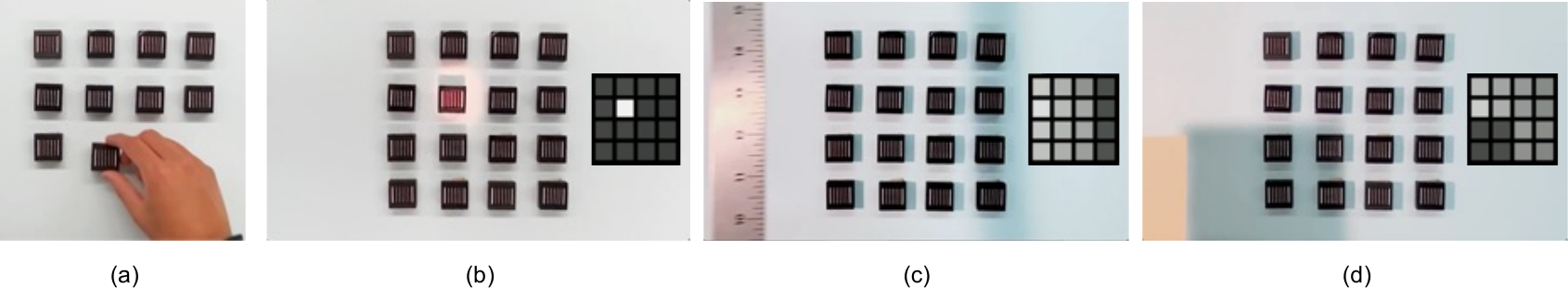}
\vspace*{-0.25in}
\caption{{\bf A self-powered pixel array.} (a) Crickets are manually arranged to emulate a $4 \times 4$ camera. Since each cricket has a unique carrier frequency \fid, chirps from a large number of crickets can be simultaneously received by the software radio and processed to obtain the light intensities measured by all the crickets. (b) A spot light is moved around over the array. The inset visual on the right shows the "image" captured by the array. In (c) and (d), shadows are cast on the array using a ruler and a piece of cardboard. Videos generated by the array are included in the supplemental video.}
\label{fig:array}
\end{figure*}
\begin{figure}[h!] 
\centering 
\includegraphics[width=0.85\linewidth]{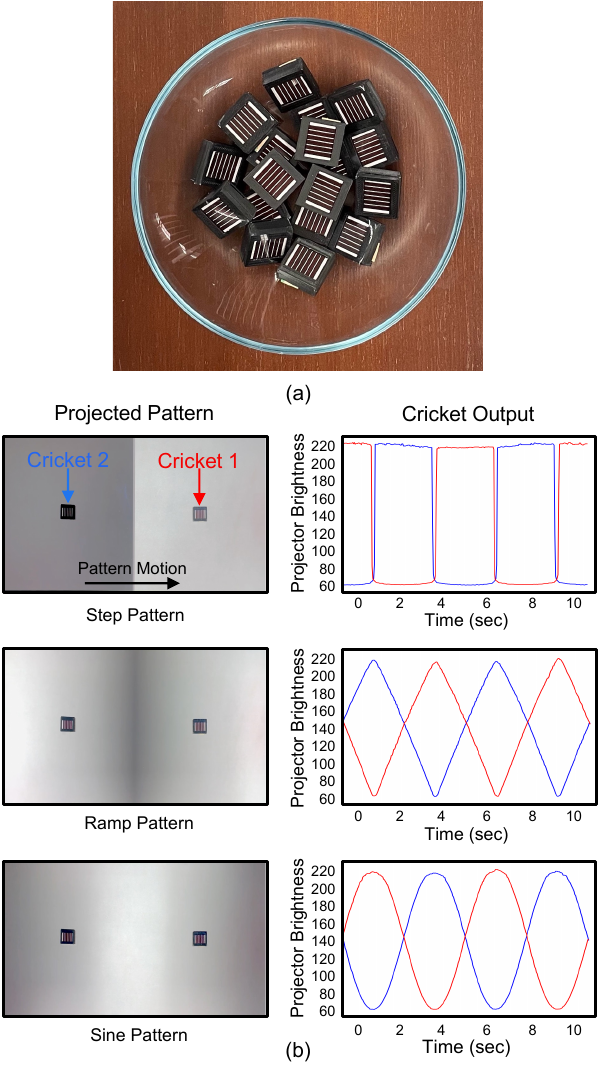}
\vspace{-0.05in}
\caption{{\bf Measuring time-varying light patterns.} (a) We have fabricated 24 crickets, each with its own chirp frequency. (b) The precision of crickets is illustrated here by projecting an illumination pattern on two crickets and sliding the pattern over the crickets as a function of time. The measured signals for the step, ramp and sine patterns demonstrate the ability of crickets to measure light with high accuracy. The supplemental video shows the brightness measured by each cricket varying with time.}
\vspace{-0.1in}
\label{fig:time-varying}
\end{figure}

\begin{figure}[h]
    \centering
    \includegraphics[width=0.94\linewidth]{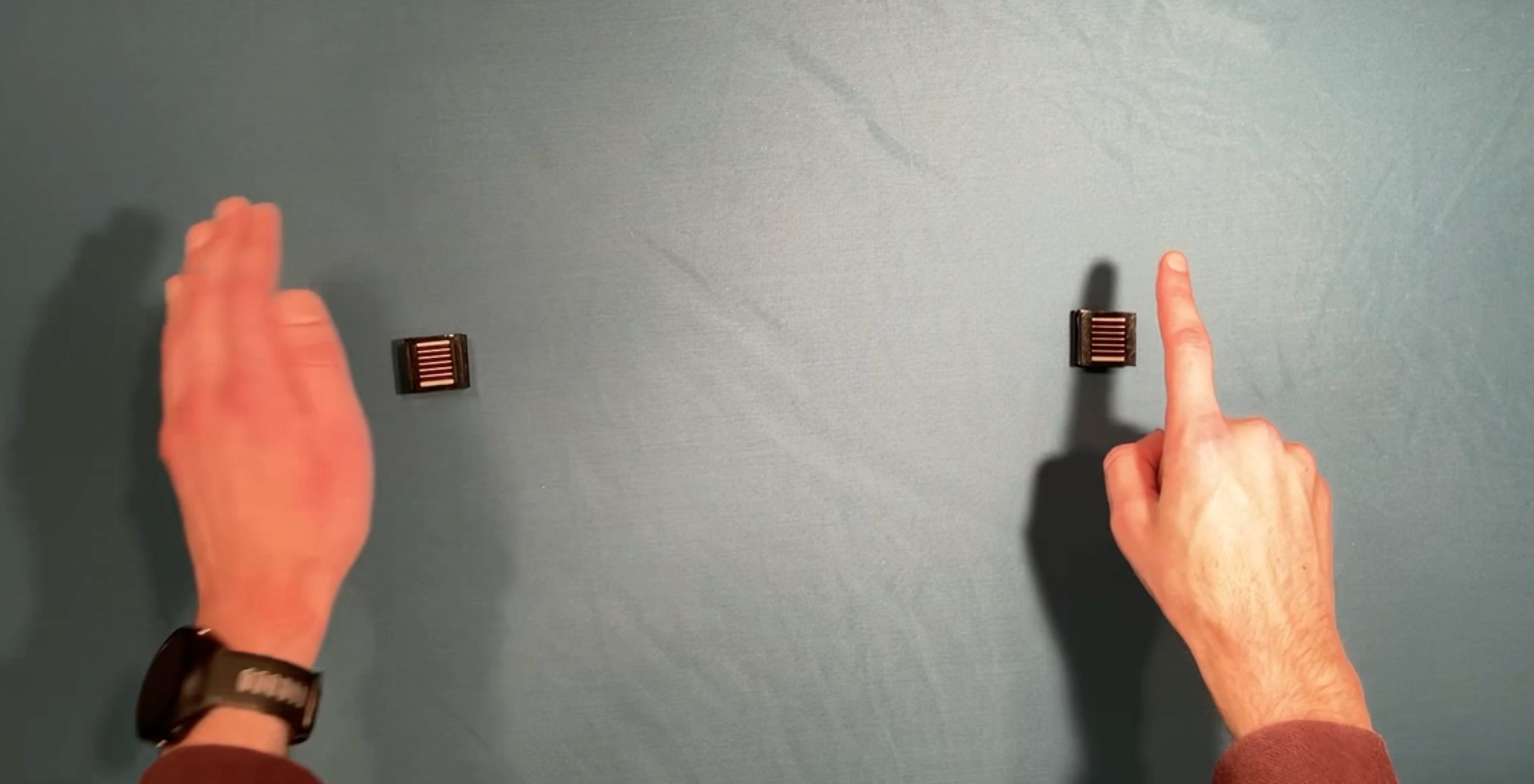}
    \caption{{\bf Luxophone.} Two crickets are used to create a musical instrument where the performer modulates the ambient light falling on the crickets to control volume and frequency. }
    \label{fig:luxophone}
    \vspace{-0.05in}
\end{figure}

\vspace{0.1in}
\noindent
{\bf Signal-to-Noise Ratio}
To quantify cricket's signal-to-noise ratio, we captured its chirps under different light levels over a duration of time. For each light level, the $N$ detected chirps were used to compute the average and variance of the light level. 
\Cref{fig:radiometry}(c) lists the peak signal-to-noise ratio (PSNR) computed for the different light levels. 
The drop in PSNR with increase in light level can be, in part, attributed to the fact that the chirps get closer as the light level increases, and hence the precision with which the time between consecutive chirps can be measured decreases.

\vspace{0.1in}
\noindent
{\bf Power:}
Crickets are designed to chirp only when they can afford to. In other words, virtually all of the harvested energy is used during a chirp, and the circuit draws minimal power between chirps. We have estimated the power consumption during a chirp to be $29 \ \unit{\mW}$, and when it is not chirping to be $2.2 \ \unit{\uW}$. Note that the number of chirps per unit time is proportional to the light level. Therefore, the power consumption, $P$, is a function of the light level, $L$ lux, and can be approximated as: $P \approx \alpha L + \beta$, where $\alpha = 116\ \unit{\nano\watt \per lux}$, and $\beta = 2.2\ \unit{\uW}$. Details of this power model are provided in the supplemental material.

\vspace{0.1in}
\noindent
{\bf Size:} As mentioned above, as long as the photovoltaic can produce sufficient voltage, the cricket will chirp. The photovoltaic we used has an active area of 
$255 \ \unit{\square \mm}$. In \cref{fig:cricket-size}, we show how the active area of the photovoltaic effects the chirp frequency \fc. In this experiment, we emulated different active areas by using a linear stage with micrometer accuracy to slide an opaque mask over the photovoltaic while it was exposed to a light intensity of $700 \ \unit{lux}$. As expected, the chirp frequency \fc\ falls with the active area. Interestingly, the cricket continues to chirp with an active area as small as $2.89 \ \unit{\square \mm}$, which is a reduction in active area by a factor of $100\times$ and corresponds to a photovoltaic with an active area of $1.7 \times 1.7 \ \unit{\square \mm}$.
\vspace{0.1in}

\Cref{fig:time-varying} demonstrates the measurement accuracy of crickets. Two crickets are placed at a distance from each other, and illumination patterns (step, ramp and sine functions) are swiped passed the crickets using a projector. The plots on the right side show the brightness measured by each cricket as a function of time. In \cref{fig:array}(a), 16 crickets are used to create a $4 \times 4$ grid of pixels. Since each cricket has a different carrier frequency \fid, the chirps from all the crickets are simultaneously detected and used to construct the image shown within each frame. The array is illuminated using a spot light in \cref{fig:array}(b), and shadows are cast on it using a ruler and a card in \cref{fig:array}(c) and \cref{fig:array}(d), respectively.

On the lighter side, we used crickets to create a musical instrument, which we refer to as a Luxophone (\cref{fig:luxophone}). Inspired by the Theremin~\shortcite{theremin_method_1928}, an instrument that uses capacitance to wirelessly sense a musician's hand motions to control the volume and frequency of a single note, the Luxophone maps the light falling on two crickets to volume and frequency. As seen in \cref{fig:luxophone}, a performer creates music by modulating the ambient light falling on the two crickets. Music generated using the luxaphone is included in the supplemental video.

\begin{figure}
         \hspace{-0.2in}
    \includegraphics[width=3.3in]{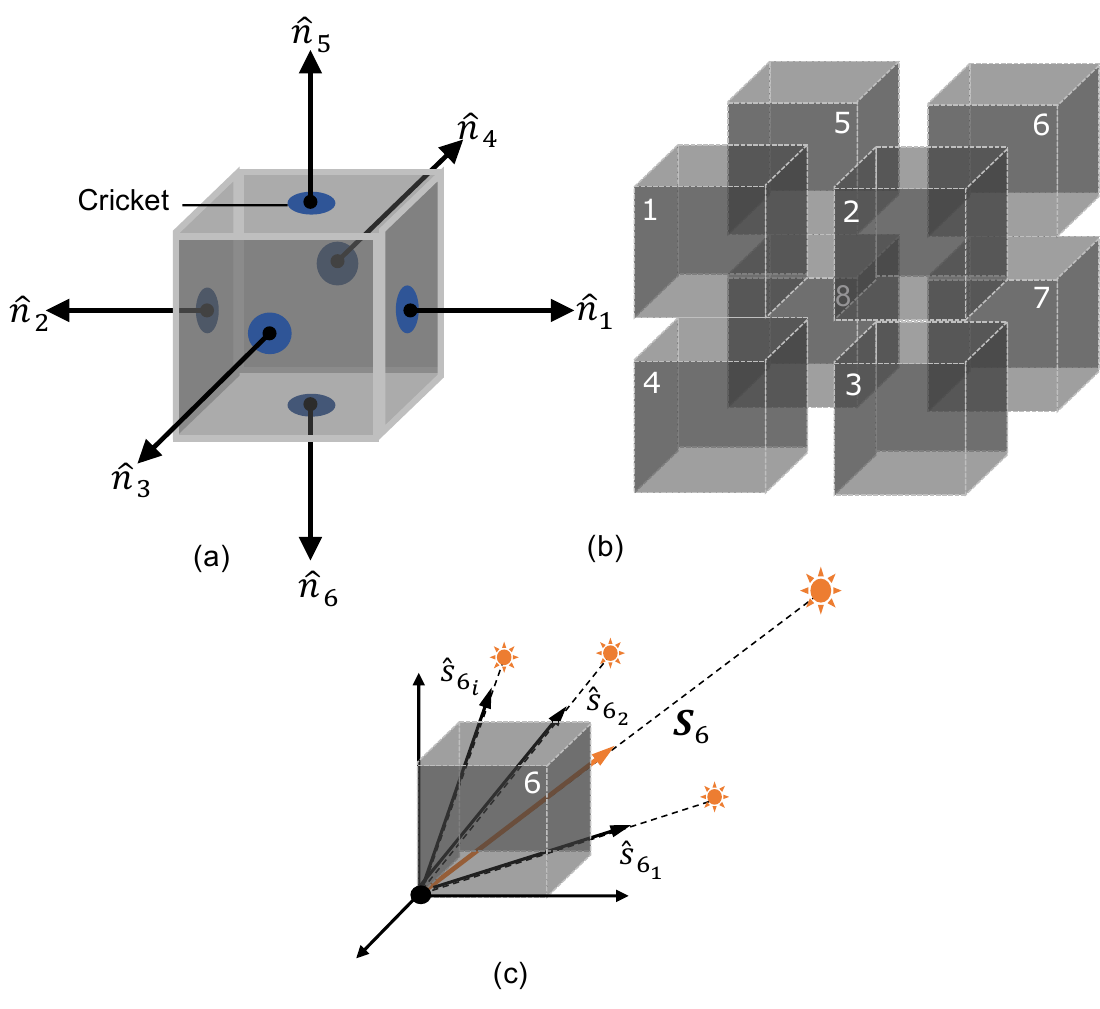}
       \vspace{-0.05in}
       \caption{{\bf Measuring the illumination centroid using a cricket cube.} (a) A cube with a cricket on each of its six faces. (b) Each cricket can only receive light from four octants. (c) All the sources within an octant can be represented with a single "resultant" source.}
    \label{fig:cube-concept}
       \vspace{-0.05in}
       \end{figure}

 \section{Cricket Cube for Source Estimation}\label{sec:cube}
 
 The estimation of the illumination of a scene has several applications in vision and graphics. Examples include surface reconstruction using shape from brightness methods, determining material properties and scene relighting. In other fields, illumination estimation can enable solar panels to track the sun, estimate weather parameters and monitor lighting in an urban setting.

If the illumination is restricted to a single point source, three crickets oriented differently are sufficient to estimate the brightness and direction of the source. Assume that the three crickets have known orientations \none, \ntwo\  and \nthree, and the unknown point source is $\Ssource = \ksource \ssource$, where \ksource \, is its brightness and \ssource \, is its direction. Let the three cricket measurements be \Ione, \Itwo\  and \Ithree. Then, we can find the point source as $\Ssource = \Nvect^{-1} \Ivect$, where $\Nvect=[\none,\ntwo,\nthree ]^{T}$ and $\Ivect = [\Ione, \Itwo, \Ithree]^{T}$. This result can be viewed as the inverse of photometric stereo \cite{woodham_photometric_1980}, where three known sources, activated in sequence, can be used to compute the surface normal of a Lambertian scene point. Note that the above point source estimation using crickets works only when the source is “visible” to all three crickets. 

Now, consider the case where the illumination is complex with an arbitrary number of unknown point sources,\footnote{Although for our purposes here we view the illumination as being comprised of a set of point sources, our results generalize to extended sources as they can be modeled as collections of point sources.} and we wish to find the “centroid” of the entire illumination field. The centroid is of particular value in applications such as solar panel tracking, where we want the panel to be tilted in the direction of maximum irradiance to maximize energy harvesting. Numerous methods have been proposed for sensing the sun direction for solar tracking~\cite{solar-review}. Note that the centroid direction could differ from the sun direction when the sky contribution is significant.

Consider the cube shown in \cref{fig:cube-concept}(a), which has a cricket on each of its six sides. Since the crickets are on a cube, their normals are related as:
\begin{equation}
\ntwo  =  -\none, \, \, \, 
\nfour  =  -\nthree, \, \, \, 
\nsix  =  -\nfive \,.
\label{eq:normal-constraint}
\end{equation}

\begin{figure}
    \hspace*{-0.1in}
    \includegraphics[width=3.25in]{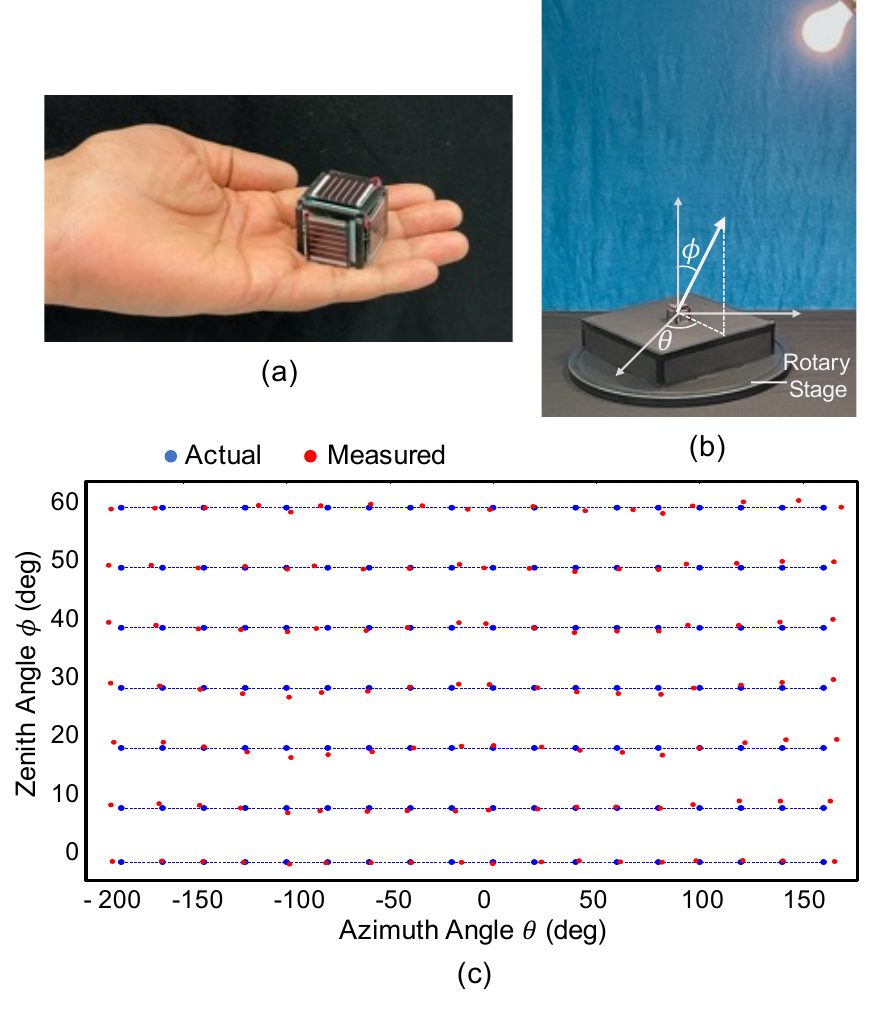}
        \vspace{-0.05in}
        \caption{{\bf Source estimation using a cricket cube.} (a) Prototype of a cricket cube. (b) Setup used to estimate the accuracy of single source estimation. The zenith angle $\phi$ and azimuthal angle $\theta$ were varied using a rotary stage and a source with adjustable height. (c) The actual (blue dots) and measured (red dots) source directions. The mean absolute errors in $\theta$ and $\phi$ were found to be $2.37^{o}$ and $0.49^{o}$, respectively.}
    \label{fig:point-source}
    \vspace{-0.1in}
\end{figure}
\begin{figure*}[t]
\centering 
\includegraphics[width=0.92\linewidth]{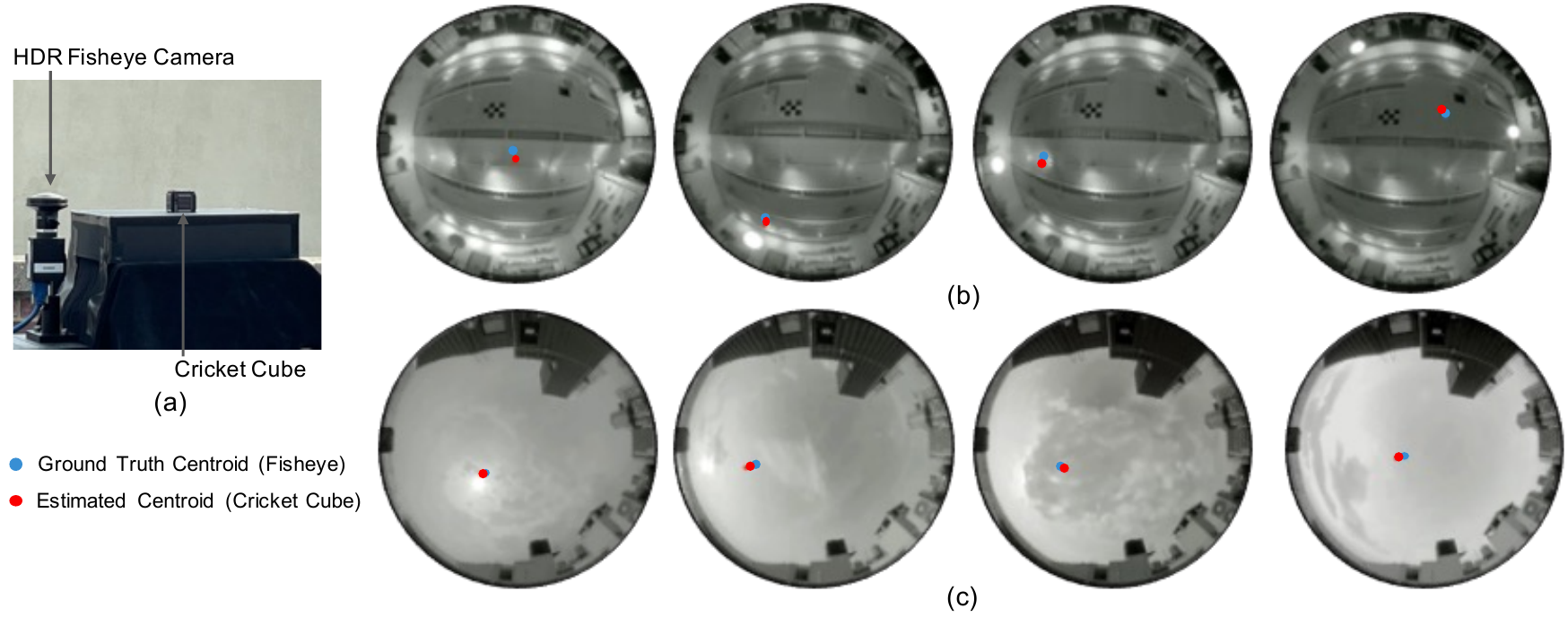}
\vspace{-0.05in}
\caption{{\bf Illumination centroid estimation using a cricket cube.} (a) For each time instant, the ground truth centroid was estimated using HDR images computed from 8 differently exposed images taken using the fisheye camera. Estimated centroids for (b) an indoor scene and (c) the sky. Note that the centroid estimates obtained from just the 6 cricket measurements produced by the cube closely track the ground truth. This can be attributed to the high precision and wide dynamic range of crickets.}
\label{fig:centroid}
\end{figure*}

Assume the illumination to be represented on a sphere and the cube to be located at the center of the sphere. To account for the fact that the sphere is only partially visible to each cricket, we partition the sphere into its 8 octants, as shown in \cref{fig:cube-concept}(b). Now, consider octant 6, which includes a number of sources with unknown brightnesses and directions, as shown in \cref{fig:cube-concept}(c). All of these sources can be represented with a single ``resultant'' source: $\Ssix = \sum_{i}^{} \ki \ssixi.$
Since different octants do not share sources, the resultant source for the entire sphere is:
\begin{equation}
\Sres = \sum_{i=1}^{8} \Si \, .
\end{equation}
Note that any given cricket can only receive light from the four octants visible to it. For instance, cricket 1 can only receive light from octants 2, 3, 6, and 7. So, we can express the light measurements from the six crickets as:
\begin{eqnarray} 
\nonumber
\Ione & = & \Stwo \none + \Sthree \none + \Ssix \none +  \Sseven \none \, ,\\ \nonumber
\Itwo & = & \Sone \ntwo + \Sfour \ntwo + \Sfive \ntwo +  \Seight \ntwo \, ,\\ \nonumber
\Ithree & = & \Sone \nthree + \Stwo \nthree + \Sthree \nthree +  \Sfour \nthree \, ,\\ 
\Ifour & = & \Sfive \nfour + \Ssix \nfour + \Sseven \nfour +  \Seight \nfour \, ,\\ \nonumber
\Ifive & = & \Sone \nfive + \Stwo \nfive + \Sfive \nfive +  \Ssix \nfive \, ,\\ \nonumber
\Isix & = & \Sthree \nsix + \Sfour \nsix + \Sseven \nsix +  \Seight \nsix \, .
\label{eq:cube-measurements}
\end{eqnarray}
From the above equations and the constraints in \cref{eq:normal-constraint}, we obtain the resultant source as:
\vspace{-0.04in}
\begin{equation}
\Sres = {\Nmatrix}^{-1} \Ivect \, ,
\end{equation}
where, $\Nmatrix = [\none, \nthree, \nfive]^{T}$ and $\Ivect = [\Ione - \Itwo, \, \Ithree - \Ifour , \, \Ifive - \Isix ]^{T}$. The direction of the centroid of the illumination is simply the direction of the resultant source: $\scent = \Sres / \| \Sres \|$.

The cricket cube we developed is shown in \cref{fig:point-source}(a). We first evaluated its ability to estimate the centroid direction $\scent$ by using a single point source, in which case $\scent$ is the direction of the source. As seen in \cref{fig:point-source}(b), a rotary stage was used to vary the azimuthal angle $\theta$, and the height of the source was varied to set the zenith angle $\phi$. The cube was able to estimate the source direction with high accuracy – the mean absolute errors in $\theta$ and $\phi$ were found to be $2.37\unit{\degree}$ and $0.49\unit{\degree}$, respectively. 

Next, we used the cube to find the illumination centroid for both indoor and outdoor scenes. To obtain ground truth, we used the fisheye camera shown in \cref{fig:centroid}(a). To capture the wide dynamic ranges of these scenes, which included table and ceiling lamps indoor and direct sunlight outdoor, for each instant in time, we captured 8 fisheye images using different exposures and computed a high dynamic range (HDR) image. The ground truth centroid (blue dot) was computed from the HDR image. In the frames shown in \cref{fig:centroid}(b) and \cref{fig:centroid}(c), the centroid estimated using the six cricket cube measurements (red dot) is very close to the ground truth in both the indoor and outdoor examples. The mean absolute error in the angle between the ground truth and cube-estimated centroids for the indoor and outdoor scenes were found to be $4.14\unit{\degree}$ and $2.92\unit{\degree}$, respectively. It is worth reiterating that we only used six measurements to estimate the centroid. This performance can be attributed to the cricket's high sensitivity and wide dynamic range.

\begin{figure*}[t]
\centering 
\includegraphics[width=0.95\linewidth]{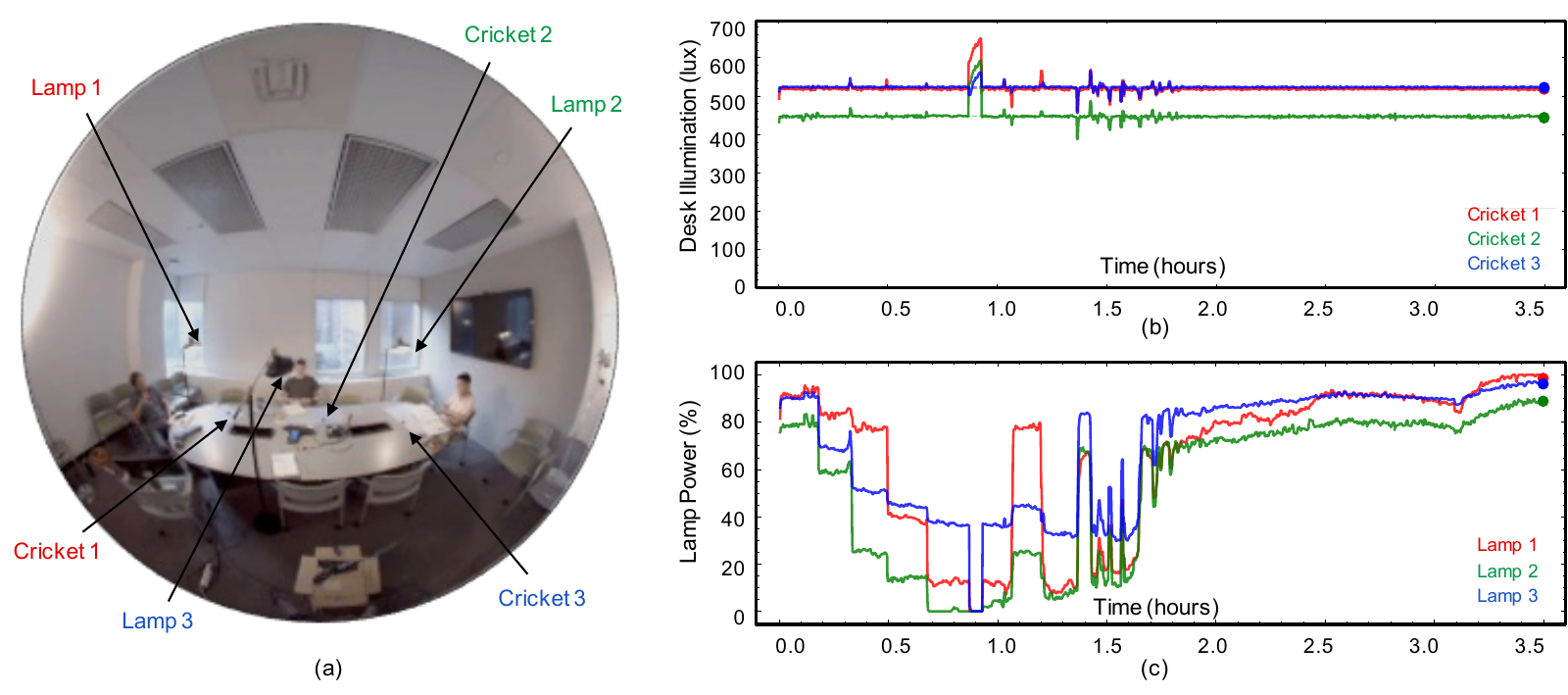}
\caption{{\bf Lighting control using crickets.} (a) Three crickets are positioned in regions where we wish the illumination to be kept constant. The lighting was controlled using three Wi-Fi enabled LED lamps. Over a span of 3.5 hours, the ambient illumination varied dramatically due to light streaming through the windows and the ceiling lights being turned on and off. (b) Despite the significant changes in ambient illumination, the lighting of the three regions remained more or less constant. (c) The lamp intensities generated by the control algorithm. In this example, lighting control led to an energy saving of $37\%$ compared to keeping the three lights at maximum brightness.}
\label{fig:lighting-control}
\end{figure*}

\vspace{0.1in}
\section{Lighting Control}\label{sec:lighting}

\vspace{0.05in}
Lighting control for energy conservation is a problem that has been extensively studied. The idea is to continuously adjust the intensities of lamps in an indoor setting (offices, factories, nurseries, etc.) based on measurements provided by light sensors that are positioned in regions where lighting needs to be controlled (desks, entrances, plants, etc.). In these settings, there is typically some ambient illumination (e.g., sunlight through windows) that varies in an unpredictable manner over time. 

Previous works in this area have explored the design of sensor networks and  control algorithms for adjusting lighting~\cite{wen_towards_2006, park_design_2007, pan_wsn-based_2008, yeh_autonomous_2010}. Pandharipande and Li~\shortcite{pandharipande_light-harvesting_2013} have developed a light-powered sensor that uses a microcontroller-based circuit to measure illumination and detect motion. 
Since these sensors require more power than crickets, they use a larger solar cell and rely on a battery to function in lower light conditions.
In comparison, crickets can measure light down to 10 lux without a battery. They are compact and can be easily placed in the scene without being conspicuous (see \cref{fig:lighting-control}(a)). Furthermore, as shown in \cref{fig:cricket-size}, the current prototype can be significantly reduced in size, 
enabling them to be inconspicuously embedded in a variety of everyday objects (furniture, devices, etc.).

Assume we have an indoor space with $n$ controllable lamps and $m$ regions where we have placed crickets. 
We represent the lamp intensities using the vector \Lamp, where each of its elements is normalized to lie between $0$ and $1$. We denote the measurements obtained from the crickets as \Meas. The direct effect of the lamp intensities on the cricket measurements  is given by the transport matrix \Trans: $\Meas = \Trans \Lamp$, where \Trans \ is an $m \times n$ matrix. We measure \Trans \ offline by eliminating all the ambient light (turning off other lights and shutting the windows), fully turning on each of the controllable lights in sequence, and recording the corresponding cricket measurements. 

The lighting model for the space can now be written as $\Meas = \Trans \Lamp + \Amb$, where \Amb \ is the unknown, time-varying ambient illumination. Let \Measdes \ be the desired cricket measurements dictated by the application, \Measnow \ be the current cricket measurements, and \Lampnow \ be the current lamp intensities. Our control algorithm finds the next (adjusted) lamp intensities \Lampnext \ so as to minimize the discrepancy between \Measdes\ and \Measnext. We assume that the ambient lighting varies slowly, and hence \Amb\ is the same at times $t$ and $t + 1$. We can then define a loss function that represents the difference between the desired and actual measurements at time $t + 1$ as:
\begin{equation}
    \loss\left(\Lampnext\right) = || \left( \Trans \Lampnext + \Amb \right) - \Measdes ||^2. \label{eq:lighting-loss-1}
\end{equation}
Using our lighting model, we can express the ambient light as: $\Amb = \Measnow - \Trans \Lampnow$. Substituting this into \cref{eq:lighting-loss-1}, we get:
\begin{equation}
    \loss\left(\Lampnext\right) = || \Trans \left( \Lampnext - \Lampnow \right) + \Measnow - \Measdes ||^2. \label{eq:lighting-loss-2}
\end{equation}
Our lighting control algorithm finds the next lamp intensities \Lampnext\ by minimizing \cref{eq:lighting-loss-2} subject to the constraint $0 \le \Lampnext \le 1$, which can be solved using quadratic programming. 

\Cref{fig:lighting-control} shows the above method applied to a conference room with windows, ceiling lights and three controllable lights, over a span of 3.5 hours. Three crickets are placed close to the three users of the conference table and their desired measurements were chosen to be $520 \ \unit{lux}$, $450 \ \unit{lux}$ and $515 \ \unit{lux}$. In this experiment, each lamp had a non-zero contribution to each of the three crickets. That is, all the elements of the transport matrix \Trans\ were non-zero.

The ambient lighting of the scene varied dramatically during the experiment (please see supplemental video). Despite these changes, as seen in \cref{fig:lighting-control}(b), the outputs of the three crickets remain more or less constant. The short deviations in the cricket measurements seen in the plot are due to the fact that some of the ambient lighting changes were abrupt. In  \cref{fig:lighting-control}(c), we show the optimal lamp intensities computed by our algorithm. As is evident, the lamp intensities had to be varied significantly over time to ensure the cricket measurements remained close to the desired values. From these plots we estimated that the total energy saved by controlling the lamp intensities, versus keeping them at maximum brightness, was $37 \%$.

\begin{figure*}[t]
  \centering 
\includegraphics[width=0.96\linewidth]{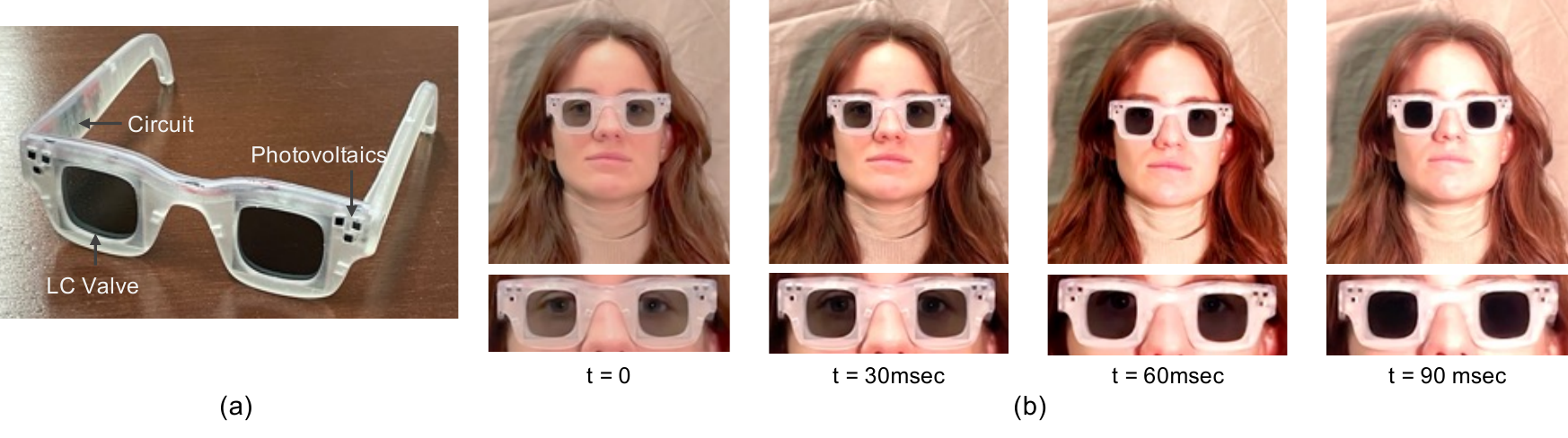}
\vspace*{-0.2in}
\caption{{\bf eTransition glasses.} (a) A modified version of the cricket circuit is used to develop sunglasses that can automatically adjust their transmittance based on the brightness of the environment. (b) The glasses can adapt to lighting changes within 100 milliseconds compared to today's widely used photochromic transition lenses that take over 30 seconds to darken and over 2 minutes to lighten.}  
\label{fig:glass-proto}
\end{figure*}

\begin{figure}[h]
  \hspace{-0.25in}  
  \centering 
    \includegraphics[width=1.0\linewidth]{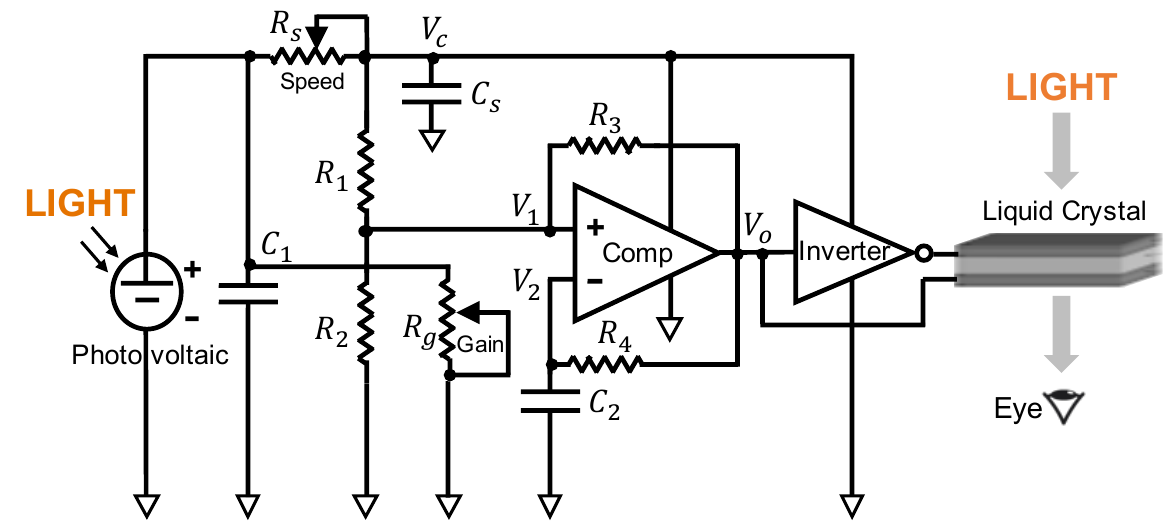}
  \vspace{-0.1in}  
  \caption{{\bf eTransition glass circuit.} 
The output \Vo\ of the comparator in \cref{fig:cricket-circuit} is a train of pulses. We modified this light-powered pulse generator so that its peak value is proportional to the incident light and its duty cycle is $50\%$. This output is used to drive a liquid crystal (LC) valve such that its transmittance is inversely proportional to the incident light. This circuit was used to develop the battery-less transition glasses shown in \cref{fig:glass-proto}(a).}
  \label{fig:glass-circuit}
  \vspace{-0.15in}
\end{figure}

\section{{\small E}Transition Glasses}\label{sec:glasses}

In our last application, we describe a pair of transition sunglasses that use liquid crystal (LC) valves that are driven by a self-powered circuit (\cref{fig:glass-circuit}) inspired by cricket. Note that the first part of the cricket circuit in \cref{fig:cricket-circuit} -- from the photovoltaic to the comparator output -- is essentially a self-powered pulse generator with a very low duty cycle. For our transition glasses, we modified cricket's pulse generator and changed the values of some of the components (see supplemental material) so that the output \Vo\ of the comparator has a duty cycle of $50\%$ and its peak voltage is proportional to the incident light. In place of the oscillator in \cref{fig:cricket-circuit}, we use an inverter that converts \Vo\ to an AC pulse train that is used to drive the LC valve. The transmittance of the LC valve is therefore inversely proportional to the incident light. The resistors \Rspeed\ and \Rgain\ are used to control the transition speed and attenuation range, respectively. In the supplemental material, we have provided a more detailed description of how the circuit works as well as the values and model numbers of the components we used in our implementation.

\Cref{fig:glass-proto}(a) shows a prototype of our ``eTransition'' glasses. The frame includes 3 miniature solar cells on each side, an LC valve for each eye, and the circuit in \cref{fig:glass-circuit}. Multiple solar cells are used in this case only because each one has a low peak voltage of $0.4 \, \unit{V}$ and a minimum of $2.0 \, \unit{V}$ is needed to drive the LC valves. \Cref{fig:glass-proto}(b) shows four video frames of a user captured while the illumination of the environment is increased by turning on a bright source. As can be seen, the transition is completed within a span of $90 \ \unit{\ms}$.

\section{Discussion}\label{sec:future}

There are several directions along which we plan to extend our work. First, we plan to miniaturize our cricket design so we can deploy them in large numbers both in the form of arrays as well as a network of light sensors. This will involve the custom design of the photovoltaic cell, the oscillator and a single integrated circuit for the remaining components. Our current implementation uses an off-the-shelf antenna. It is well known that custom designing the antenna for a specific frequency and form-factor can significantly enhance its efficiency. Antenna design is an involved task, but we believe it is a worthwhile investment since we would like our miniaturized cricket to have a longer transmission range than our current prototype. 

Second, while we have demonstrated the use of cricket cubes for measuring point source direction and centroid estimation for arbitrary illumination, we plan to investigate how a small number of crickets can be used to create untethered outward-looking and inward-looking light probes that can provide richer descriptions of illumination. 

Finally, we believe that the concept of a cricket can be generalized to a light-powered environmental sensor that can measure not only light but other parameters such as temperature, pressure and humidity. There are a variety of components, including passive ones such as thermistors, that can be incorporated into our current design. The key challenge is to ensure that the additional measurements are faithfully encoded within the chirps transmitted by the cricket.

\begin{acks}

This work was supported by the Office of Naval Research (ONR) under award number N00014-23-1-2096-P00003. The authors are grateful to Behzad Kamgar-Parsi of ONR for his support and encouragement during the course of this work. The authors thank Vijay Modi for his feedback on the use of crickets for solar tracking, Harish Krishnaswamy for his input regarding FCC compliance, Mia Chiquier for her help with the experiments and Eray Baykal for 3D printing the case of the eTransition glasses. 

\end{acks}

\bibliographystyle{ACM-Reference-Format}
\bibliography{Cricket}

\setcounter{section}{0}
\renewcommand\thesection{\Alph{section}}
\renewcommand\thesubsection{\thesection.\arabic{subsection}}

\clearpage

\section{Cricket Circuit Details}
\begin{figure}[t]
	\centering
	\includegraphics[width=\linewidth]{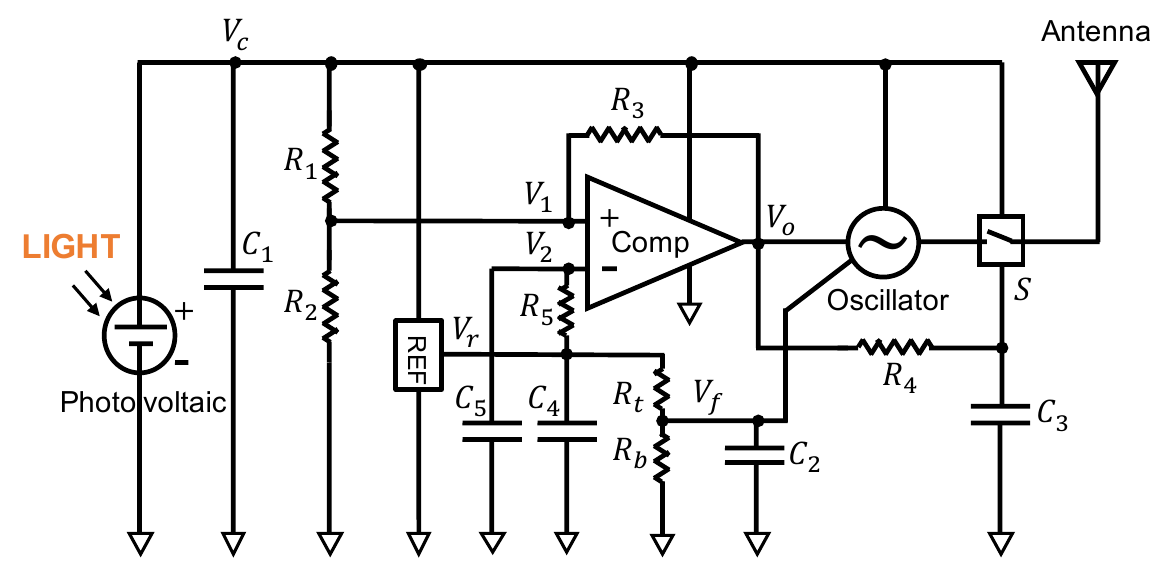}
	\caption{{\bf Cricket circuit.}}
	\label{fig:supp-cricket-circuit}
\end{figure}

Here we provide a detailed explanation of the cricket circuit in \cref{fig:supp-cricket-circuit}. When the photovoltaic cell is exposed to light, it charges capacitor \Cone, and voltage \Vc\ begins to rise. At some voltage, the comparator comes to life. At that point, the reference (regulated) voltage \Vr\ is smaller than the voltage \Vone\ and hence the output \Vo\ of the comparator is $0$.  Since \Vc\ continues to rise, at some point, \Vone\ will exceed \Vtwo, and the comparator output \Vo\ goes from $0$ to \Vc.  This activates the oscillator, which puts out an RF frequency (in the GHz range). The frequency \fid\ of the oscillator output, which is the identifier of the cricket, is determined by the voltage \Vf, which, in turn, is determined by the resistors \Rt\ and \Rb\ and the capacitor \Ctwo.  In our implementation, \fid\ is preset by selecting the resistor \Rb. 

Although the oscillator is now active, it is not yet connected to the antenna due to the switch \Sw. The closing of this switch is delayed by \Rfour\ and \Cthree. This delay is introduced to allow the preset voltage \Vf\ to stabilize and ensure the frequency applied to the antenna is precise and stable. In the current implementation, this delay has been set to roughly $10 \ \unit{\micro s}$. After the delay, the oscillator is connected to the antenna, which transmits an RF chirp for about $30 \ \unit{\micro s}$. This chirp duration is limited solely by the fact that the oscillator is, by far, the highest power consumer in the circuit. Hence, while the oscillator is active, \Vc\ falls rapidly until \Vone\ goes lower than \Vtwo, the comparator output goes to $0,$ and the oscillator shuts down. At this point, \Cone\ begins to recharge and \Vc\ rises again. 

Our current implementation uses off-the-shelf passive (resistors and capacitors) and active (comparator, oscillator and switch) components. The values and model numbers of the components we used in our prototype are given in \cref{tab:supp-cricket-component-values}.
The resistor \Rthree\ introduces hysteresis in the comparator output. It can be lowered to increase the length of the chirp. In our case, however, we want the chirp to be very short and hence we have used a large value ($100 \ \unit{\Mohm}$).

\section{Power Model Details}
We have estimated the power consumption of a cricket from the typical current drain of each of its components. In \cref{tab:supp-power-computation}, we list the current drain for each component when the cricket when the cricket is chirping and sleeping (not chirping). When the cricket is chirping, the oscillator consumes $10 \ \unit{\mA}$ during its $40 \ \unit{\micro \second}$ chirp. When the cricket is sleeping, the active components and voltage dividers consume $767 \ \unit{\nA}$. 

The power consumed by each component is also function of the voltage. We will assume a constant voltage $\Vc = 2.9 \ \unit{\V}$ to compute the power consumption from the currents in \cref{tab:supp-power-computation}.\footnote{Technically, \Vc\ is not a constant but a sawtooth wave shown in Fig. 4 of the main paper. To approximate the power consumption, we conservatively use $\Vc = 2.9 \ \unit{\V}$, the threshold voltage at which the cricket chirps.} This is a valid choice for all components, including those powered by \Vr, because the voltage reference is a linear regulator. Using this voltage, the power consumed while chirping and sleeping is:
\begin{equation}
    P_{chirping} = 29.0 \ \unit{\mW}, \ \ P_{sleeping} = 2.23 \ \unit{\uW}.
\end{equation}
A typical cricket chirps roughly $0.1$ chirps per lux, per second, and each chirp lasts $40 \ \unit{\micro \second}$. From these facts, we can compute the time spent chirping and sleeping as a function of the light level, $L \ \unit{lux}$, in a time interval $T$ seconds as:
\begin{equation}
    T_{chirping} = 4LT \cdot 10^{-6} \ \unit{\second}, \ \ T_{sleeping} = T - T_{chirping} \ \unit{\second}.
\end{equation}
Now, the average power consumption is given by:
\begin{equation}
    P = \frac{T_{chirping} \cdot P_{chirping} + T_{sleeping} \cdot P_{sleeping}}{T} \ \unit{\W}.
\end{equation}
We can rewrite $P$ as a function of the light level, $L \ \unit{lux}$, as $P = \alpha L + \beta$, where $\alpha = 116 \ \unit{\nano\watt \per lux}$ and $\beta = 2.23 \ \unit{\uW}$.

\begin{table}[t]
	\caption{\textbf{List of components in the circket circuit.}}
	\label{tab:supp-cricket-component-values}
	\centering
	\begin{tabular}{lcc}
		\toprule
		Component & Value \\
		\midrule
		\Cone & $3.3 \ \unit{\uF}$ \\
		\Ctwo & $10 \ \unit{\uF}$ \\
		\Cthree & $220 \ \unit{\pF}$ \\
		\Cfour & $0.1 \ \unit{\uF}$ \\
		\Cfive & $1 \ \unit{\uF}$ \\
		\Rone & $6.2 \ \unit{\Mohm}$ \\
		\Rtwo & $10 \ \unit{\Mohm}$ \\
		\Rthree & $100 \ \unit{\Mohm}$ \\
		\Rfour & $100 \ \unit{\kohm}$ \\
		\Rfive & $1 \ \unit{\Mohm}$ \\
		\Rt & $4.02 \ \unit{\Mohm}$ \\
		\Rb & $10 - 100 \ \unit{\Mohm}$ (adjustable) \\
		Photovoltaic & Panasonic AM-5610CAw-DGK-T \\
		Voltage Reference & ABLIC Inc. S-1318D18-M5T1U4 \\
		Comparator & Texas Instruments TLV3691 \\
		Oscillator & Analog Devices MAX2752 \\
		Switch & Analog Devices ADG 901 \\
		Antenna & Taoglas FXP 29 ($2.04-2.10 \ \unit{\GHz}$) \\
		\bottomrule
	\end{tabular}
\end{table}

\begin{table}[ttt]
    \caption{\textbf{Current drain of components in the cricket circuit.}}
    \label{tab:supp-power-computation}
    \centering
    \begin{tabular}{lcc}
        \toprule
         Component & Chirping (nA) & Sleeping (nA) \\
         \midrule
         Oscillator & $10^7$ & 200 \\
         Voltage divider,\\\Rone\ and \Rtwo & 181 & 181 \\
         Voltage divider,\\\Rt\ and \Rb & 116 & 116 \\
         Switch & 100 & 100 \\
         Voltage reference & 95 & 95 \\
         Comparator & 75 & 75 \\
         \midrule
         Total & $10^7$ & 767 \\
         
         \bottomrule
    \end{tabular}
\end{table}
\section{{\small E}Transition Glass Circuit Details}
\begin{figure}[t]
	\centering
	\includegraphics[width=\linewidth]{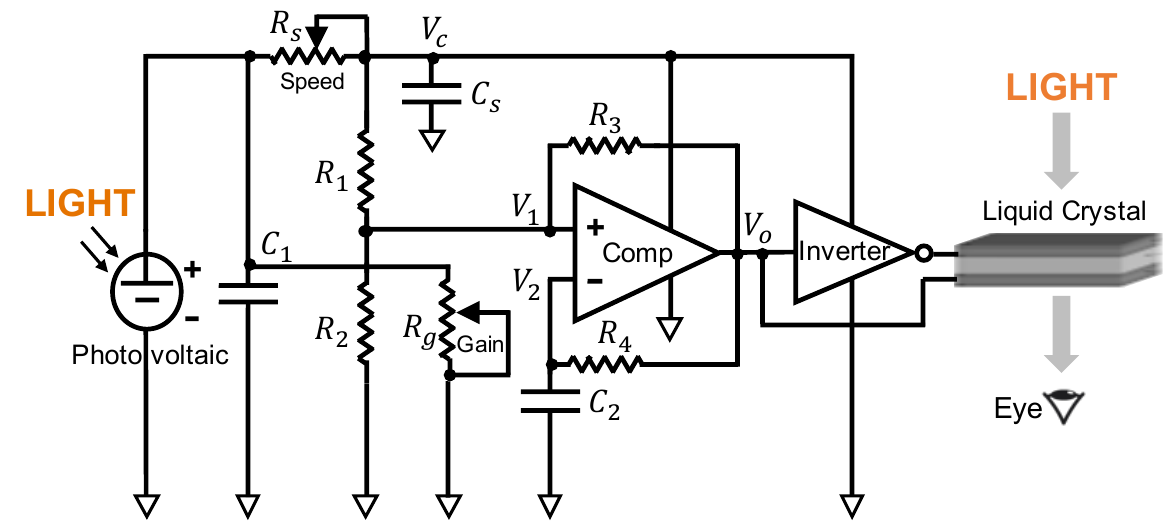}
	\caption{{\bf eTransition glass circuit.}}
	\label{fig:supp-glass-circuit}
\end{figure}

Here we detail the eTransition glass circuit in \cref{fig:supp-glass-circuit}, a self-powered circuit inspired by cricket that drives a liquid crystal (LC) light valve. As mentioned in the main paper, the first part of the cricket circuit in \cref{fig:supp-cricket-circuit} -- from the photovoltaic to the comparator output -- is essentially a self-powered pulse generator with a very low duty cycle. In place of the oscillator, we use an inverter that converts \Vo\ to an AC pulse train in order to drive the LC valve. The transmittance of the LC valve is therefore inversely proportional to the incident light. 

As with the cricket, when the photovoltaic is exposed to light, the voltage \Vc\ begins to rise. Initially, the comparator output \Vo\ is $0$. When \Vone\ exceeds \Vtwo, \Vo\ jumps to \Vc. At this point, \Vtwo\ begins to rise due to \Rfour\ and \Ctwo. At some point, \Vtwo\ exceeds \Vone, and the comparator output \Vo\ drops to $0$. The comparator output \Vo\ is therefore a square wave and its duty cycle is determined by \Rone, \Rtwo\ and \Rthree. We have chosen these resistors such that the duty cycle is 50\%. The frequency of the square wave is determined by \Rfour\ and \Ctwo. Most importantly, the peak value of the square wave is proportional to the intensity of light falling on the photovoltaic. \Cref{tab:supp-glasses-component-values} lists the component values used in this circuit.

We now have \Vo\ varying between $0$ and \Vc. However, we need an AC voltage which includes a reverse in polarity. This is achieved by using the inverter. Both the output of the comparator and the output of the inverter have a push-pull configuration. These outputs are used to, in effect, create an H-bridge, which results in the LC valve seeing an AC voltage that flips between (\Vo, 0) and (0, \Vo), where \Vo\ is proportional to the incident light intensity. 

\begin{table}[t]
	\caption{\textbf{List of components in the eTransition glass circuit.}}
	\label{tab:supp-glasses-component-values}
	\centering
	\begin{tabular}{lcc}
		\toprule
		Component & Value \\
		\midrule
		\Cone & $3.3 \ \unit{\uF}$ \\
		\Ctwo & $100 \ \unit{\pF}$ \\
		\Cspeed & $0.1 \ \unit{uF}$ \\
		\Rone & $10 \ \unit{\Mohm}$ \\
		\Rtwo & $10 \ \unit{\Mohm}$ \\
		\Rthree & $10 \ \unit{\Mohm}$ \\
		\Rfour & $10 \ \unit{\Mohm}$ \\
		\Rgain & $100 \ \unit{\Mohm}$ \\
		\Rspeed & $1 \ \unit{\kohm}$ \\
		Photovoltaic ($\times$6) & Vishay BPW34 \\
		Comparator & Texas Instruments TLV3691 \\
		Inverter & Texas Instruments SN74AUP1G04 \\
		LC Light Valve ($\times$2) & Adafruit PID 3627 \\
		\bottomrule
	\end{tabular}
\end{table}

\end{document}